\documentclass[letterpaper]{article}
\usepackage{uai2019}
\usepackage[margin=1in]{geometry}

\usepackage{times}
\PassOptionsToPackage{sort, numbers, compress}{natbib}

\title{Learning Markov Chain in Unordered Dataset}

\author{Yao-Hung Hubert Tsai$^\dagger$, Han Zhao$^\dagger$, Nebojsa Jojic$^\ddagger$, Ruslan Salakhutdinov$^\dagger$\\
$^\dagger$Machine Learning Department, Carnegie Mellon University\\
$^\ddagger$Microsoft Research
}

\usepackage{hyperref}
\usepackage{changes}
\usepackage{url}
\usepackage{booktabs}
\usepackage{amssymb}
\usepackage{enumitem}
\usepackage{graphicx}
\usepackage{color}
\usepackage{subcaption}
\usepackage{hyperref}
\usepackage{mathtools}
\usepackage{amsmath}
\usepackage{amssymb}
\usepackage{amsthm}
\usepackage{amsfonts}       
\usepackage{nicefrac}       
\usepackage{microtype}      
\usepackage{graphicx}
\usepackage{tabularx}
\usepackage{multirow}
\usepackage{algorithm}
\usepackage{algorithmic}
\usepackage{float}
\usepackage{wrapfig}
\usepackage{listings}
\usepackage{times}
\usepackage{thmtools}
\usepackage{thm-restate}
\usepackage[round,sort,comma,super,authoryear]{natbib}

\theoremstyle{definition}
\newtheorem{definition}{Definition}[section]

\newtheorem{theorem}{Theorem}[section]

\newtheorem{proposition}{Proposition}[section]
\newtheorem{corollary}{Corollary}[section]

\renewcommand{\algorithmiccomment}[1]{\bgroup\hfill//~#1\egroup}

\newcommand{\RR}{\mathbb{R}}

\DeclareMathOperator*{\argmin}{arg\,min}
\DeclareMathOperator*{\argmax}{arg\,max}

\newcommand{\defeq}{\vcentcolon=}

\definecolor{dkgreen}{rgb}{0,0.6,0}
\definecolor{gray}{rgb}{0.5,0.5,0.5}
\definecolor{mauve}{rgb}{0.58,0,0.82}
\newcommand{\model}{OrderNet~}
\newcommand{\sspace}{\mathcal{S}}
\newcommand{\transition}{\mathcal{T}}
\newcommand{\transitional}[3]{\mathcal{T}_{#3}(#1~|~#2)}

\newcommand{\Q}{\mathbb{Q}}

\begin{document}

\maketitle

\begin{abstract}
The assumption that data samples are independently identically distributed is the backbone of many learning algorithms. Nevertheless, datasets often exhibit rich structure in practice, and we argue that there exist some unknown order within the data instances. In this technical report, we introduce \model that can be used to extract the order of data instances in an unsupervised way. By assuming that the instances are sampled from a Markov chain, our goal is to learn the transitional operator of the underlying Markov chain, as well as the order by maximizing the generation probability under all possible data permutations. Specifically, we use neural network as a compact and soft lookup table to approximate the possibly huge, but discrete transition matrix. This strategy allows us to amortize the space complexity with a single model. Furthermore, this simple and compact representation also provides a short description to the dataset and generalizes to unseen instances as well. To ensure that the learned Markov chain is ergodic, we propose a greedy batch-wise permutation scheme that allows fast training. Empirically, we show that \model is able to discover an order among data instances. We also extend the proposed \model to one-shot recognition task and demonstrate favorable results.
\vspace{-4mm}
\end{abstract}

\section{Introduction}
\vspace{-1mm}
\label{sec:intro}
Recent advances in deep neural networks offer a great potential for both supervised learning, e.g., classification and regression~\citep{glorot2011domain,krizhevsky2012imagenet,xu2015show,finn2017deep} and unsupervised learning, e.g., dimensionality reduction and density estimation~\citep{hinton2006reducing,vincent2010stacked,chen2012marginalized,kingma2013auto,goodfellow2014generative,rasmus2015semi}. In domains where data exhibit natural and explicit sequential structures, sequential prediction with recurrent neural networks and its variants are abundant as well, i.e., machine translation~\citep{bahdanau2014neural,sutskever2014sequence,wu2016google}, caption generation~\citep{vinyals2015show}, short-text conversation~\citep{shang2015neural}, to name a few. However, despite the wide application of deep models in various domains, it is still unclear whether we can find implicit sequential structures in data without explicit supervision? We argue that such sequential order often exists even when dealing with the data that are naturally thought of as being $i.i.d.$ sampled from a common, perhaps complex, distribution. For example, consider a dataset consisting of the joint locations on the body of the same person taken on different days. The $i.i.d.$ assumption is justified since postures of a person taken on different days are likely unrelated. However, we can often arrange the data instances so that the joints follow an articulated motion or a set of motions in a way that makes each pose highly predictable given the previous one. Although this arrangement is individually dependent, for example, ballerina's pose might obey different dynamics than the pose of a tennis player, the simultaneous inference on the pose dynamics can lead to a robust model that explains the correlations among joints. As another example, if we reshuffle the frames of a video clip, the data can now be modeled under the $i.i.d.$ assumption. Nevertheless, reconstructing the order leads to an alternative model where transitions between the frames are easier to fit. 
\begin{figure*}[htb]
	\centering
	\includegraphics[width=1.0\linewidth]{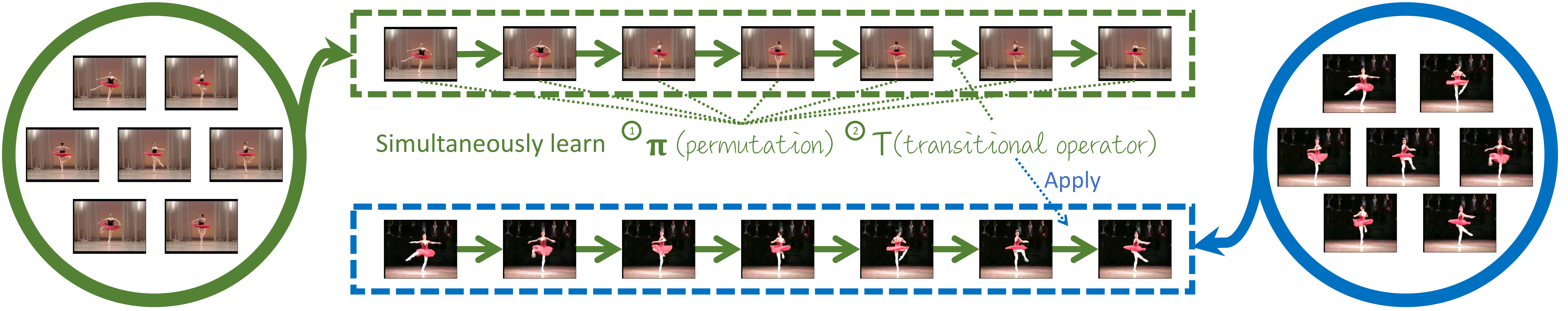}
	\caption{ Illustration of the proposed model \model. The data are assumed to be generated from a Markov chain, and \model learns the permutation and the transition matrix simultaneously. The figure illustrates that we first train \model for the data in green circles (without knowing the data order) to recover the data permutation. Then, we apply the trained transition operator to the data in blue circles to infer data permutation.}
	\label{fig:illus}
\end{figure*}

In this technical report, we investigate the following question: 
\begin{quote}
\emph{Given an unordered dataset where instances may exhibit some implicit order, can we find this order in an unsupervised way?}
\end{quote}
One naive approach is to perform sorting based on a predefined distance metric, e.g., the Euclidean distance between image pixel values. However, the distance metrics have to be predefined differently according to distinct types/characteristics of the datasets at hand. A proper distance metric for one domain may not be a good one for other domains. For instance, the $p$-distance is a good measure for DNA/RNA sequences \citep{nei2000molecular}, while it does not characterize the semantic distances between images. We argue that the key lies in the discovery of proper distance metric automatically and adaptively. Despite all kinds of interesting applications, to the best of our knowledge, we are the first to study this problem.

To approach this problem, we model the data instances by assuming that they are generated from a Markov chain. We propose to simultaneously train the transition operator and find the best order by a joint optimization over the parameter space as well as all possible permutations. We term our model \model. One of the key ideas in the design of \model is to use neural networks as a soft lookup table to approximate the possibly huge but discrete transition matrix, which allows \model to amortize the space complexity using a unified model. Furthermore, due to the smoothness of the function approximator implemented by neural networks, the transitional operator of \model can also generalize on unseen but similar data instances. To ensure the Markov chain learned by \model is ergodic, we also propose a greedy batch-wise permutation scheme that allows fast training. Fig. \ref{fig:illus} illustrates our proposed model.

As an application, we further extend our model to one-shot recognition tasks, where only one labeled data is given per category in the target domain. Most of the current work in this area focuses on learning a specific distance metric \citep{koch2015siamese,vinyals2016matching,snell2017prototypical} or category-separation metric \citep{ravi2016optimization,finn2017model} for the given data. During the inference phase, one would then compute either the smallest distance or highest class prediction score between the support and query instances. Alternatively, from a generative modeling perspective, we can first generate samples from the learned Markov chain, starting at each of the support instances, evaluate the query instance under the generated samples, and assign the label with the highest probability. Empirically, we demonstrate that \model is able to discover implicit orders among data instances while performing comparably with many of the current state-of-the-art methods on one-shot recognition tasks.

\section{The Model}
\label{sec:model}
We begin with introducing the setup of the problem and notations used in this technical report.  Let $\{s_{i}\}_{i=1}^n$ denote our training data which are assumed to be generated from an unknown, fully observable, discrete time, homogeneous Markov chain $\langle \sspace, \transition\rangle$, where $\sspace$ is the state space and $\transition$ is the transitional operator over $\sspace$, i.e., $\transition: \sspace\to\Delta(\sspace)$, a distribution over $\sspace$. In this work we assume the initial distribution is uniform over $\sspace$.\footnote{This is just for notational convenience as we can estimate the initial distribution over a set of trajectories.} Let $d = |\sspace|$ be the size of the state space. We denote the underlying data order to be a permutation over $[n]$: $\pi = \{\pi_t\}_{t=1}^n$, where $\pi_t$ represents the index of the instance generated at the $t$-th step of the Markov chain. Our goal is to jointly learn the transitional operator of the Markov chain as well as the most probable order of the generation process.

\subsection{Parametrized Transitional Operator via Neural Networks}
\label{subsec:NN}
In classic work of learning Markov chain~\citep{sutton1998reinforcement}, the transitional operator is usually estimated as a discrete table that maintains counts of transitions between pairs of states. However, in practice, when the state space is large, we cannot often afford to maintain the tabular transition matrix directly, which takes up to $O(d^2)$ space. For example, if the state refers to a binary image $I \in \{0,1\}^p$, then $d = 2^{2p}$, which is computationally intractable to maintain explicitly, even for $p$ of moderate size. 

In this technical report, instead of maintaining the potentially huge discrete table, our workaround is to parametrize the transitional operator $\transition$ as $\transitional{\cdot}{\cdot}{\theta}: \sspace \times \sspace\to\RR_+$ with learnable parameter $\theta$. Being universal function approximators~\citep{hornik1989multilayer}, neural networks could be used to efficiently approximate discrete structures which led to the recent success of deep reinforcement learning \citep{mnih2013playing,guo2014deep,oh2015action,silver2017mastering}. In our case, we use a neural network to approximate the discrete tabular transition matrix. The advantages are two-fold: first, it significantly reduces the space complexity by amortizing the space required by each separate state into a unified model. Since all the states share the same model as the transition operator, there is no need to store the transition vector for each separate state explicitly. Second, neural networks allow better generalization for the transition probabilities across states. The reason is that, in most real-world applications, states, represented as feature vectors, are not independent from each other. As a result, the differentiable approximation to a discrete structure has the additional smoothness property, which allows the transition operator to generalize on unseen states. 

More specifically, $\transitional{s'}{s}{\theta}$ takes two states $s$ and $s'$ as its input and returns the corresponding (unnormalized) transition probability from $s$ to $s'$ as follows:
\begin{equation}
\Pr(s'~|~s) = \frac{1}{Z_s}\transitional{s'}{s}{\theta}, \quad Z_s = \sum_{s'\in\sspace}\transitional{s'}{s}{\theta}
\label{equ:trans}
\end{equation}
where $Z_s$ is the normalization constant of state $s$. Note that one can consider each discrete transition matrix as a lookup table, which takes a pair of indices of the states and return the corresponding transition probability; for example, we can use $s$ and $s'$ to locate the corresponding row and column of the table and read out its probability. From this perspective, the neural network works as a soft lookup table that outputs the unnormalized transition probability given two states (features). We will describe how to implement the transitional operator for both continuous and discrete state spaces in more detail in Sec.~\ref{subsec:param_oper}.

\subsection{A Greedy Approximation of the Optimal Order}
\label{subsec:order}
Given an unordered data set $\{s_i\}_{i=1}^n$, the problem of joint learning the transitional operator and the generation order can be formulated as the following maximum likelihood estimation problem:
\begin{equation}
\text{maximize}_{\theta,\pi}\quad \sum_{t=2}^n \log\transitional{s_{\pi_t}}{s_{\pi_{t-1}}}{\theta}
\label{equ:opt}
\end{equation}
where the permutation $\pi$ is chosen from all possible permutations over $[n]$. It is worth pointing out that the optimization problem \eqref{equ:opt} is intrinsically hard, as even if the true transitional operator was given, it would still be computationally intractable to find the optimal order that maximizes \eqref{equ:opt}, which turns out to be an NP-hard problem. 
\begin{proposition}
\label{prop:hard}
Given the transitional operator $\mathcal{T}$ and a set of instances $\{s_{i}\}_{i=1}^n$, finding the optimal generation order $\pi$ is NP-hard.
\end{proposition}

To prove the proposition, we construct a polynomial time mapping reduction from a variant of the traveling salesman problem (\textsf{TSP}). To proceed, we first formally define the traveling salesman path problem (\textsf{TSPP}) and our optimal order problem in Markov chain (\textsf{ORDER}).

	\begin{definition}[\textsf{TSPP}] 
	INSTANCE: A weighted undirected graph $G = (V, E, w)$ and a constant $C$. QUESTION: Is there a Hamiltonian path in $G$ whose weight is at most $C$?
	\end{definition}

	\begin{definition}[\textsf{ORDER}]
	INSTANCE: A Markov chain $M = \langle \sspace, \transition\rangle$, a sequence of states $X$ and a constant $p$, where $\sspace\defeq [n]$ is the set of states, $\transition\in\Q_+^{n\times n}$ is the transitional matrix. QUESTION: Is there a permutation $\pi$ for $X$ under which $M$ generates $\pi(X)$ with probability at least $p$?
	\end{definition}

	\begin{theorem}[\citep{garey2002computers}]
	\textsf{TSPP} is \textsf{NP}-complete.
	\end{theorem}

	\begin{theorem}
	$\textsf{TSPP}\leq_m^p \textsf{ORDER}$.
	\end{theorem}
	\begin{proof}
	Given an instance of \textsf{TSPP}, $\langle G = (V, E, w), C\rangle$, we shall construct an instance of \textsf{ORDER}, $\langle M = \langle \sspace, \transition\rangle, X, p\rangle$ in polynomial time such that the answer to the latter is ``yes'' if and only if the answer to the first is also ``yes''. 

	Given $\langle G = (V, E, w), C\rangle$, $n = |V|$, define $\Delta\defeq \max_{i\in[n]}\sum_{j=1}^n \exp(-w(e_{ij}))$. With $\Delta$, we construct a Markov chain $M$ where the state set is $S = V$ and the transitional matrix is defined as:
	\begin{equation*}
	T_{ij} \defeq \exp(-w(e_{ij})) /\Delta, \forall e_{ij}\in E
	\end{equation*}
	and
	\begin{equation*}
	T_{ii}\defeq 1- \sum_{j\neq i}T_{ij}
	\end{equation*}
	Set the initial distribution of $M$ to be uniform. Choose $X = \sspace = [n]$ and $p = \exp(-C)/n\Delta^n$. 

	Now we show $\langle G = (V, E, w), C\rangle\in \textsf{TSPP}\iff \langle M = \langle V, T\rangle, \exp(-C)/n\Delta^n\rangle\in \textsf{ORDER}$. If $\langle G = (V, E, w), C\rangle\in \textsf{TSPP}$, then there exists a Hamiltonian path over $V$ such that the weight of the path $\leq C$. In other words, there exists a permutation $\pi$ over $V$ such that $\sum_{i=2}^n w(e_{\pi(i-1), \pi(i)}) \leq C$, but this in turn implies:
	\begin{align*}
	&\Pr(\pi(X); M) = \frac{1}{n}\prod_{i=2}^n \Pr(\pi(X_i)\mid \pi(X_{i-1}))\\
	&= \frac{1}{n}\cdot\frac{\exp(\sum_{i=2}^n -w(e_{\pi(i-1), \pi(i)}))}{\Delta^n}\geq \frac{\exp(-C)}{n\Delta^n} = p	
	\end{align*}
	which shows $\langle M = \langle V, T\rangle, \exp(-C)/n\Delta^n\rangle\in \textsf{ORDER}$. The other direction is exactly the same, i.e., if we know that $\Pr(\pi(X); M) \geq \exp(-C)/n\Delta^n = p$, then we know that there is a Hamiltonian path in $G$ with weight $\leq C$.
	\end{proof}

	\begin{corollary}
	\textsf{ORDER} is \textsf{NP}-complete.
	\end{corollary}
	\begin{proof}
	We have proved that $\textsf{TSPP}\leq_m^p \textsf{ORDER}$. Along with the fact that $\textsf{TSPP}$ is NP-complete, this shows $\textsf{ORDER}$ is NP-hard. Clearly, $\textsf{ORDER}\in\textsf{NP}$ as well, because a non-deterministic machine can first guess a permutation $\pi$ and verify that $\Pr(\pi(X); M) \geq p$, which can be done in polynomial time in both $n$ and $|X|$. This finishes our proof that \textsf{ORDER} is \textsf{NP}-complete. 
	\end{proof}

\begin{corollary}
Given the transitional operator $\transition$ and a set of instances $\{s_i\}_{i=1}^n$, finding the optimal generation order $\pi$ is NP-hard.
\end{corollary}
\begin{proof}
Since the decision version of the optimization problem is NP-complete, it immediately follows that the optimization problem is NP-hard.
\end{proof}

Given the hardness in finding the optimal order, we can only hope for an approximate solution. To this end, we propose a greedy algorithm to find an order given the current estimation of the transition operator. The algorithm is rather straightforward: it first enumerates all possible data $s_i$ as the first state, and then given state at time step $t-1$, it greedily finds the state at time step $t$ as the one which has the largest transition probability from the current state under the current transitional operator. The final approximate order is then defined to be the maximum of all these $n$ orders. We list the pseudocode in Alg.~\ref{alg:greedy}. This algorithm has time complexity $O(n^3)$. To reduce the complexity for large $n$, we can uniformly at random sample a data to be the one generated at the first time step, and again using greedy search to compute an approximate order. This helps to reduce the complexity to $O(n^2)$. Note that more advanced heuristics exist for finding the approximate order $\pi$, e.g., the genetic algorithm, simulated annealing, tabu search, to name a few. 
\begin{algorithm}[t!]
\small
\centering
\caption{ Greedy Approximate Order}
\label{alg:greedy}
\begin{algorithmic}[1]
\REQUIRE    Input data $\{s_i\}_{i=1}^n$ and transitional operator $\transitional{s'}{s}{\theta}$.
\FOR        {$i = 1$ to $n$}
	\STATE	$\pi_1^{(i)}\leftarrow i$
    \FOR {$j = 2$ to $n$}
    	\STATE	$\pi_j^{(i)} \leftarrow \max_{k\not\in \pi^{(i)}_{1:j-1}} \transitional{s_k}{s_{\pi^{(i)}_{j-1}}}{\theta}$\COMMENT{greedy selection}
    \ENDFOR
\ENDFOR
\STATE	$\hat{\pi}\leftarrow\argmax_{i\in[n]} \sum_{t=2}^n \log\transitional{s_{\pi^{(i)}_{t-1}}}{s_{\pi^{(i)}_t}}{\theta}$
\end{algorithmic}
\end{algorithm}

Using Alg.~\ref{alg:greedy} as a subroutine, at a high level, the overall optimization algorithm can be understood as an instance of coordinate ascent, where we alternatively optimize over the transitional operator $\transition_\theta$ and the approximate order of the data $\hat{\pi}$. Given $\hat{\pi}$, we can optimize $\theta$ by gradient ascent. In what follows, we introduce a batch-wise permutation training scheme to further reduce the complexity of Alg.~\ref{alg:greedy} when $n$ is large, so that is scales linearly with $n$.

\subsection{Batch-Wise Permutation Training}
\label{subsec:bwper}
The $O(n^3)$ computation to find the approximate order in Alg.~\ref{alg:greedy} can still be expensive when the size of the data is large. In this section we provide batch-wise permutation training to avoid this issue. The idea is to partition the original training set into batches with size $b$ and perform greedy approximate order on each batch. Assuming $b\ll n$ is a constant, the effective time complexity becomes: $O(b^3)\cdot n/b = O(nb^2)$, which is linear in $n$.

However, since training data are partitioned into chunks, the learned transition operator is not guaranteed to have nonzero transition probabilities between data from different chunks. In other words, the learned transition operator does not necessarily induce an ergodic Markov chain due to the isolated chuck of states, which corresponds to disconnected components of the transition graph. To avoid this problem, we use a simple strategy to enforce that some samples overlap between the consecutive batches. We show the pseudocode in Alg.~\ref{alg:gdmc}, 
where  $b$ means the batch size, $\gamma$ is the learning rate and $b_0 < b$ is the number of overlapping states between consecutive batches. 
\begin{algorithm}[t!]
\small
\centering
\caption{ Batch-Wise Permutation Training}
\label{alg:gdmc}
\begin{algorithmic}[1]
\REQUIRE	$\{s_i\}_{i=1}^n$, $b_{0}$, $b$, $t$, $\gamma$
\STATE	Initialize $\theta^{(0)}$, $\{x^{(0)}_i\}_{i=1}^b$
\FOR {$k = 1$ to $\infty$}
\IF {$k \equiv 1 (\mathrm{mod}\,\, t)$}
\STATE Sample $\{x^{(k)}_{i}\}_{i=1}^{b-b_{0}} \sim \{s_{i}\}_{i=1}^n$
\STATE $\{x^{(k)}_{i}\}_{i=1}^{b} \leftarrow \{x^{(k)}_{i}\}_{i=1}^{b-b_{0}}\cup \{x^{(k-1)}_{i}\}_{i=1}^{b_{0}}$\COMMENT{construct overlapping states}
\ENDIF
\STATE Compute $\hat{\pi}^{(k)}$ using the Greedy Approximate Order (Alg.~\ref{alg:greedy}) on $\{x^{(k)}_{i}\}_{i=1}^{b}$
\STATE Compute $\nabla^{(k-1)}_{\theta}\log\Pr(\{x_i\}_{i=1}^b;\theta^{(k-1)}, \hat{\pi}^{(k)})$
\STATE	$\theta^{(k)} = \theta^{(k-1)} + \gamma \nabla^{(k-1)}_{\theta}\log\Pr(\{x_i\}_{i=1}^b;\theta^{(k-1)}, \hat{\pi}^{(k)})$\COMMENT{update of transitional operator}
\ENDFOR
\end{algorithmic}
\end{algorithm}

\subsection{Explicit Transitional Operator}
\label{subsec:param_oper}
We now provide a detailed description on how to implement and parametrize the transitional operator. The proposed transitional operator works for both discrete and continuous state space, which we discuss separately. Note that different from previous work~\citep{song2017nice}, where the only requirement for the transitional operator is a good support of efficient sampling, in our setting, we need to form explicit transitional operator that allows us to evaluate the probability density or mass of transitions. We call such parametrized transitional operators as explicit transitional operators, to distinguish them from previous implicit ones. 

\begin{figure*}[t!]
\centering
\includegraphics[width=0.75\textwidth]{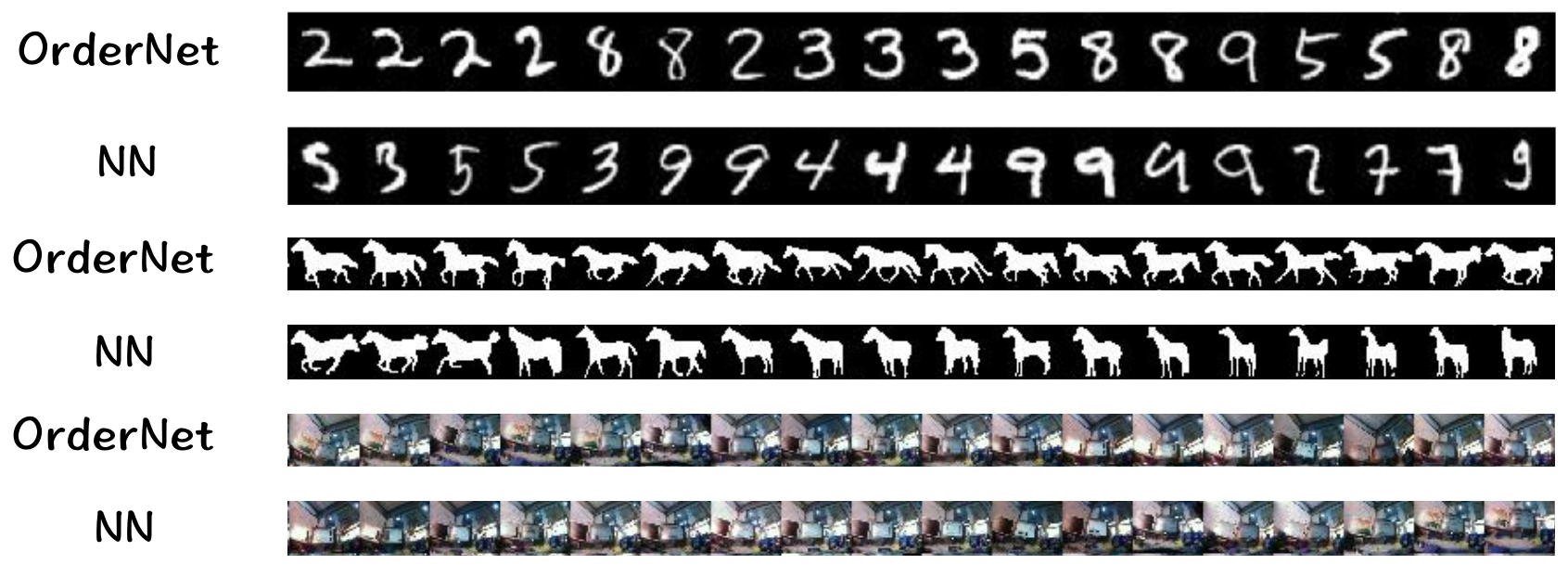}
\caption{ The implicit order observed from \model and the order implied from Nearest Neighbor sorting on MNIST, Horse, and MSR-SenseCam datasets.}
\label{fig:implicit}
\end{figure*}

\textbf{Continuous state space}. Let state $s\in\RR^p$ be a $p$-dimensional feature vector. In this case we assume that $s'\sim \mathcal{N}(m_{\theta}(s), v_\theta(s))$, where both the mean and the variance of the normal distribution are implemented with neural networks. Similar ideas on parametrizing the mean and variance functions of the normal distribution have recently been widely applied, e.g., variational autoencoder (VAE)~\citep{kingma2013auto} and its variants, etc. Under this model the evaluation of the transitional operator becomes:
\begin{equation}
\begin{split}
&\log\transitional{s'}{s}{\theta} \propto \\
&-\frac{1}{2}(s' - m_\theta(s))^Tv^{-1}_\theta(s)(s' - m_\theta(s)) - \log\det(v_\theta(s))
\end{split}
\label{equ:ceval}
\end{equation}
In practice, due to the large $O(p^2)$ space complexity of learning a full covariance matrix, we further assume that the covariance is diagonal. This allows us to further simplify the last term \eqref{equ:ceval} as $\mathbf{1}_p^T\log v_\theta(s)$.

\textbf{Discrete state space}. Let state $s\in\{0, 1\}^p$ be a $p$-dimensional 0-1 vector.\footnote{This is just for notational convenience, and it can be generalized to any discrete distribution.} Without any assumption, the distribution has a support of size $O(2^p)$, which is intractable to maintain exactly. To avoid this issue, again, we adopt the conditional independence assumption for $\transitional{s'}{s}{\theta}$, i.e.,
\begin{equation}
\begin{split}
&\log\transitional{s'}{s}{\theta} = \log\prod_{j=1}^{p}\Pr(s'_j~|~s;\theta)\defeq \\
& \sum_{j=1}^ps'_j\cdot\log f_{\theta, j}(s) + (1 - s'_j)\cdot \log(1 - f_{\theta, j}(s))
\end{split}
\label{equ:deval}
\end{equation}
where $f_{\theta, j}(\cdot)$ is the $j$-th network with sigmoid output so that we can guarantee $f_{\theta, j}(\cdot) \in (0, 1)$, representing the probability $\Pr(s'_j = 1~|~s; \theta)$. In practice, the networks $\{f_{\theta, j}\}_{j=1}^p$ share all the feature transformation layers, except the last layer where each $f_{\theta, j}$ has its own sigmoid layer. Such design not only allows potential feature transfer/multitask learning among $p$ output variables, but also improves generalization by substantially reducing model size. Interestingly, the proposed operator shares many connections with the classic restricted Boltzmann machine (RBM)~\citep{hinton2006reducing}, where both models assume conditional independence at the output layer. On ther other hand, RBM is a shallow log-linear model, whereas the proposed operator allows flexible design with various deep architectures~\citep{nair2010rectified}, and the only requirement is that the output layer is sigmoidal.

As a concluding remark of this section, we note that a common assumption in designing the explicit transitional operators for both continuous and discrete state space models is that the probability density/mass decomposes in the output layer. Such assumption is introduced mainly due to the intractable space requirement on maintaining the explicit density/mass for optimization in \eqref{equ:opt}. While such assumption is strong, we empirically find in our experiments the proposed models are still powerful in discovering orders from datasets, due to the deep and flexible architectures used as feature transformation.

\section{Experiments}
\label{sec:exp}
In this section, we first qualitatively evaluate \model on MNIST~\citep{lecun1990handwritten}, Horse~\citep{borenstein2002class}, and MSR-SenseCam~\citep{NIPS2010_4092} datasets (in which data were assumed to be i.i.d.) to discover their implicit orders. Next, we will use UCF-CIL-Action dataset~\citep{shen2008view} (with order information) to quantitatively evaluate \model and its generalization ability. Last, we apply our \model on miniImageNet~\citep{vinyals2016matching,ravi2016optimization} dataset for one-shot image recognition task. 

\begin{figure*}[t!]
\centering
\includegraphics[width=0.73\textwidth]{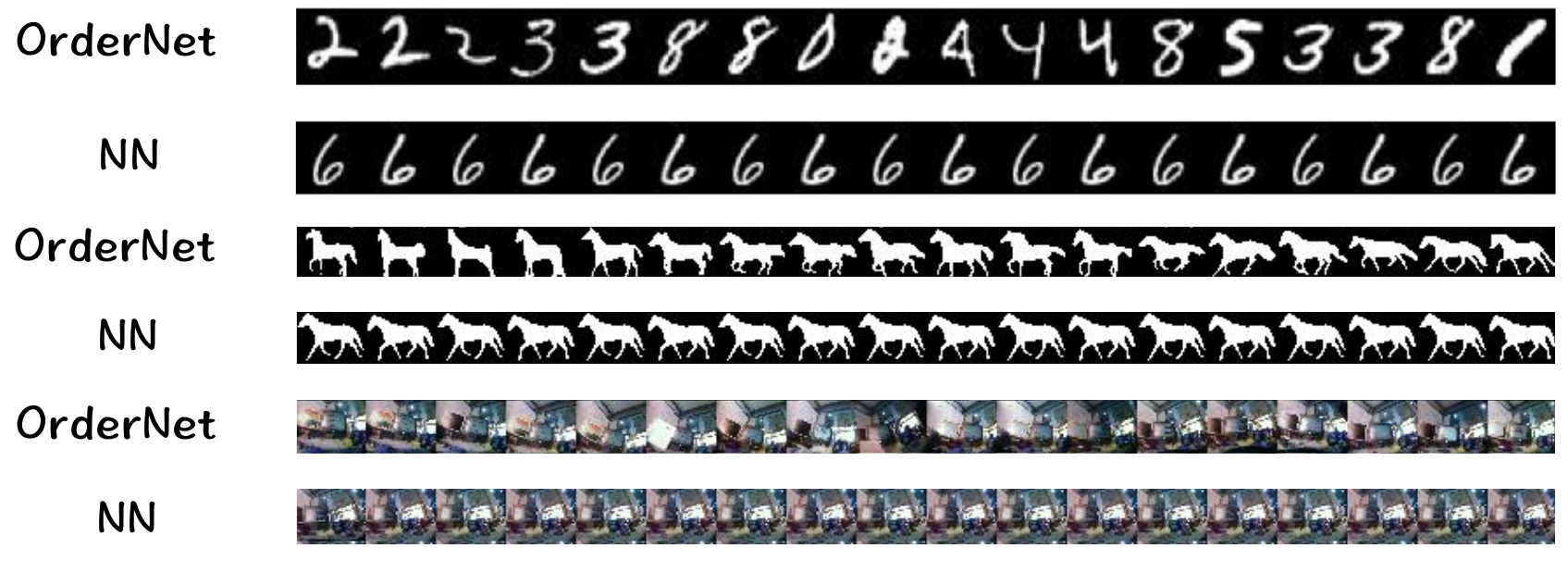}
\caption{ Image propagation for \model and Nearest Neighbor search on MNIST, Horse, and MSR-SenseCam datasets.}
\label{fig:image_prop}
\end{figure*}

\subsection{Discovering Implicit Order}
\label{subsec:ordering}
\textbf{MNIST}. MNIST  is a well-studied dataset that contains 60,000 training examples. Each example is a digit image with size 28x28. We rescale the pixel values to $[0,1]$. Note that since MNIST contains a large number of instances, we perform the ordering in a randomly sampled batch to demonstrate our results.

\textbf{Horse}. The Horse dataset consists of $328$ horse images collected from the Internet. Each horse is centered in a 30x40 image. For the preprocessing, the author applied foreground-background segmentation and set the pixel value to $1$ and $0$ for object and background, respectively. Examples are show in the supplement. 

\textbf{MSR-SenseCam}. MSR-SenseCam is a dataset consisting of images taken by SenseCam wearable camera. It contains $45$ classes with approximately $150$ images per class. Each image has size 480x640. We resize each image into 224x224 and extract the feature from VGG-19 network~\citep{simonyan2014very}. In this dataset, we consider only office category, which has $362$ images.

\subsubsection{Implicit Orders}
\label{subsubsec:implicit}
We apply Alg.~\ref{alg:gdmc} to train \model. When the training converges, we plot the images following the recovered permutation $\hat{\pi}$. Note that $\hat{\pi}$ can be seen as the implicit order suggested by \model. For comparison, we also plot the images following nearest neighbor (NN) sorting using Euclidean distances. Hyper-parameters and network architectures for parameterizing the transitional operators are specified in the supplement. The results are shown in Fig.~\ref{fig:implicit}. We first observe that consecutive frames in the order extracted by the \model have visually high autocorrelation, implying that \model can discover some implicit order for these dataset. On the other hand, it is hard to qualitatively compare the results with Nearest Neighbor's. To address this problem, we perform another evaluation in the following subsection and also perform quantitative analysis in Sec.~\ref{subsec:recover}. 

\subsubsection{Image Propagation}
\label{subsubsec:data_gen}
To verify that our proposed model can automatically learn order, we first describe an evaluation mtric, which we term as image propagation, defined as follows. Given an image $s$, the propagated image 
\begin{equation*}
s' = \argmax_{\hat{s} \in \{s_i\}_{i=1}^n \setminus s}\transitional{\hat{s}}{s}{\theta},
\end{equation*}
i.e., the most probable next image given by the transitional operator. Similar to Sec.~\ref{subsubsec:implicit}, we exploit NN search using Euclidean distance for comparison. Precisely, 
\begin{equation*}
s'_{NN} = \argmin_{\hat{s} \in \{s_i\}_{i=1}^n \setminus s}  ||s - \hat{s}||.
\end{equation*}
Fig.~\ref{fig:image_prop} illustrates the image propagation of \model and NN search. We can see that NN search would stick between two similar images. On the other hand, \model shows a series of images with consecutive actions. This implies the distinction between the discriminative (sampling by a fixed distance metric) and the generative (sampling through the transition operator in a Markov chain) model.
\subsection{Recovering Orders in Ordered Datasets}
\label{subsec:recover}
\textbf{UCF-CIL-Action}. UCF-CIL-Action~\citep{shen2008view} is a dataset containing different action videos by various subjects, distinct cameras, and diverse viewpoints. In addition to videos, the dataset also provides $11$ tracking points (presented in $x$ and $y$ axis) for head, right shoulder, right elbow, right hand, left shoulder, left elbow, left hand, right knee, right foot, left knee, and left foot. We represent a frame by concatenating $11$ tracking points at the end of pre-extracted features from VGG-19 network. In this dataset, we select ballet fouette actions to evaluate \model. Details about this experiment, including hyperparameters, model architecture, and additional experiments on the tennis serve actions, are all provided in the supplement. To quantitatively study the quality of the recovered permutation, we use Kendall Tau-b correlation score~\citep{kendall1946advanced} to measure the distance between two permutations, which outputs the value from $-1$ to $1$. The larger the Kendal Tau-b, the more similar the two permutations are. If Kendal Tau-b value is $1$, two permutations are exactly the same. In the following, we evaluate \model under two settings: (1) train \model and evaluate the order for the same subject, and (2) train \model from one subject but evaluate on the different subjects. To simulate the groundtruth ordering, we took a video sequence and then reshuffle its frames, and use \model / NN to reconstruct the order. The results are reported for random $20$ trials with mean and standard deviation.
\begin{figure*}[t!]
\centering
\includegraphics[width=1.0\linewidth]{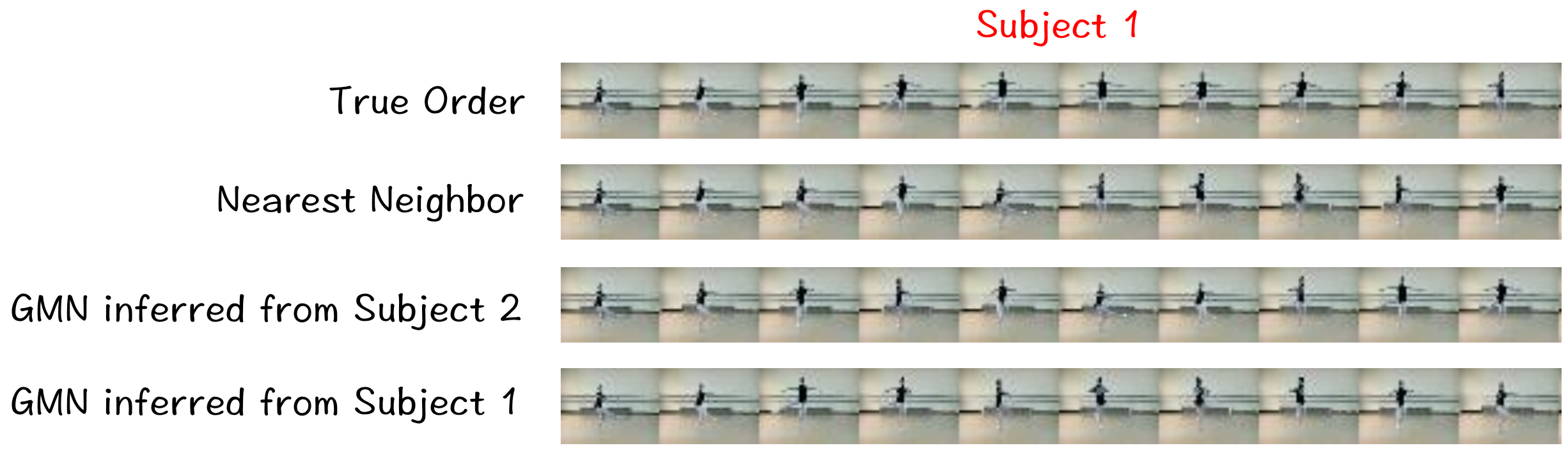}
\caption{ {\em Ballet fouette} action for $\mathsf{subject 1}$ with : (a) true order, (b) order recovered from Nearest Neighbor, (c) order recovered from \model trained on different subject, and (d) order recovered from \model trained on the same subject. Only partial orders are shown.}
\label{fig:recovery}
\end{figure*}

\begin{table*}[t!]
\centering
\caption{ Kendall Tau-b score between the true order and the recovered orders from \model/ NN with Euclidean distance. \model are trained on different subjects with ballet fouette actions.}
\label{tbl:recovery}
\begin{tabular}{|c||cccc||c|}\hline
\multirow{2}{*}{\begin{tabular}[c]{@{}c@{}}Applied To\end{tabular}} & \multicolumn{4}{c||}{Trained \model On}                                                          & \multirow{2}{*}{$\mathsf{NN}$} \\ 
                                                                      & $\mathsf{Subject\,\,1}$                    & $\mathsf{Subject\,\,2}$                   & $\mathsf{Subject\,\,3}$                    & $\mathsf{Subject\,\,4}$                    &                     \\ \hline \hline
$\mathsf{Subject\,\,1}$                                                                     & \textbf{0.364 $\pm$ 0.058} & 0.183 $\pm$ 0.033         & 0.247 $\pm$ 0.052          & 0.200 $\pm$ 0.065          & 0.137               \\ 
$\mathsf{Subject\,\,2}$                                                                     & 0.062 $\pm$ 0.091           & \textbf{0.085 $\pm$ 0.038} & 0.079 $\pm$ 0.035           & 0.074 $\pm$ 0.052           & -0.217              \\ 
$\mathsf{Subject\,\,3}$                                                                     & -0.026 $\pm$ 0.072          & -0.086 $\pm$ 0.066         & \textbf{0.191 $\pm$ 0.055} & -0.092 $\pm$ 0.083          & -0.122              \\ 
$\mathsf{Subject\,\,4}$                                                                     & 0.274 $\pm$ 0.056          &  0.288 $\pm$ 0.032         & 0.304 $\pm$ 0.025          & \textbf{0.355 $\pm$ 0.025} & 0.292               \\ \hline
\end{tabular}
\end{table*}

In Table~\ref{tbl:recovery}, we report the Kendall Tau-b values between the true order and the recovered order from \model/ NN. Qualitative results are shown in Fig.~\ref{fig:recovery}. First of all, we observe that \model can recover more accurate orders than NN according to higher Kendall Tau-b values. The most accurate order can be recovered when the \model was trained on the same subject for evaluation (diagonal in Table~\ref{tbl:recovery}). Next, we examine the generalization ability of \model. In most of the cases, when we apply the trained \model to different subjects, we can achieve higher Kendall Tau-b values comparing to NN, with only two exceptions. This result implies that \model can generalize from one subject to another.

\subsection{One-Shot Recognition}
\label{subsec:one_shot}
\textbf{MiniImageNet}. MiniImageNet is a benchmark dataset designed for evaluation of one-shot learning~\citep{vinyals2016matching,ravi2016optimization}. Being a subset of ImageNet \citep{russakovsky2015imagenet}, it contains $100$ classes and each class has $600$ images. Each image is downsampled to size 84x84. As suggested in~\citep{ravi2016optimization}, the dataset is divided into three parts: $64$ classes for training, $16$ classes for validation, and $20$ classes for testing. 

Same as~\citep{ravi2016optimization}, in this experiment we consider the $5-$way $1-$shot problem. That is, from testing classes, we sample $5$ classes with each class containing $1$ labeled example. The labeled examples are referred as support instances. Then, we randomly sample $500$ unlabeled query examples in these $5$ classes for evaluation. In more detail, we train \model on training classes and then apply it to testing classes. For each training episode, we sample $1$ class from the training classes and let $\{s_i\}_{i=1}^n$ be all the data from this class. We consider $3,000$ training episodes. On the other hand, for each testing episode, we apply \model to generate a chain from each support instance: 
\begin{equation*}
\tilde{s}_1^c \sim \transitional{\cdot}{s_0^c}{\theta}, \ldots, \tilde{s}_k^c \sim \transitional{\cdot}{\tilde{s}_{k-1}^c}{\theta},
\end{equation*}
where $s_0^c$ is the support instance belonging to class $c$ and $\tilde{s}^c$ is the generated samples from the Markov chain. Next, we fit each query example into each chain by computing the average approximating log-likelihood. Namely, the log-probability for generating the query sample $s_q$ in the chain of class $c$ is 
\begin{equation*}
\log\Pr(s_q|c) \defeq (\mathrm{log}\transitional{s_q}{s_0^c}{\theta} +  \sum_{i=1}^{k}\log\transitional{s_q}{\tilde{s}_i^c}{\theta} ) / (k+1).
\end{equation*} 
In a generative viewpoint, the predicted class $\hat{c}$ for $s_q$ is determined by
\begin{equation*}
\hat{c} = \argmax_{c}\log\Pr(s_q|c).
\end{equation*} We repeat this procedure for $10,000$ times and report the average with $95\%$ confidence intervals in Table~\ref{tbl:one_shot}.

For fair comparisons, we use the same architecture specified in \citep{ravi2016optimization} to extract $1600$-dimensional features. We pretrain the architecture using standard softmax regression on image-label pairs in training and validation classes. The architecture consists of $4$ blocks. Each block comprises a CNN layer with $64$ $3$x$3$ convolutional filters, Batch Normalization \citep{ioffe2015batch} layer, ReLU activation, and $2$x$2$ Max-Pooling layer. Then, we train \model based on these $1,600$ dimensional features. More details can be found in the supplement.
\begin{table*}[t!]
\centering
\caption{5-way 1-shot recognition task for miniImageNet. We report the mean in $10,000$ episodes with $95\%$ confidence interval.}
\label{tbl:one_shot}
\begin{tabular}{lccc}\toprule
\multicolumn{1}{l}{Model}               & Accuracy    \\\midrule 
Meta-Learner LSTM \citep{ravi2016optimization}                                          & 43.44$\pm$0.77 \\
Model-Agnostic Meta-Learning \citep{finn2017model}                          & 48.70$\pm$1.84 \\

Meta-SGD  \citep{li2017meta}                                               & 50.47$\pm$1.87 \\
\midrule
Nearest Neighbor with Cosine Distance                              & 41.08$\pm$0.70 \\
Matching Networks FCE  \citep{vinyals2016matching}                                              & 43.56$\pm$0.84 \\
Siamese    \citep{koch2015siamese}                                                & 48.42$\pm$0.79 \\
mAP-Direct Loss Minimization  \citep{triantafillou2017few}                               & 41.64$\pm$0.78 \\
mAP-Structural Support Vector Machine \citep{triantafillou2017few}                          & 47.89$\pm$0.78 \\
Prototypical Networks \citep{snell2017prototypical}                                  & 49.42$\pm$0.78 \\
\midrule
\model                         & 45.36$\pm$0.94 \\ \bottomrule
\end{tabular}
\end{table*}

\subsubsection{Results}
We compare \model and the related approaches in Table~\ref{tbl:one_shot}, in which we consider only the approaches~\citep{ravi2016optimization,finn2017deep,li2017meta,vinyals2016matching,koch2015siamese,triantafillou2017few,snell2017prototypical} using same architecture design (i.e., 4 layers CNN) in~\citep{ravi2016optimization}. 
First, we compare \model with meta-learning based methods. The best result is reported by Meta-SGD~\citep{li2017meta} with $50.47\pm 1.87$. Although \model suffers from the performance drop, it requires a much less computational budget. The reason is that the meta-learning approaches~\citep{ravi2016optimization,finn2017model,munkhdalai2017meta,li2017meta,mishra2017meta} rely on huge networks to manage complicated intersections between meta and base learners, while parameters for \model exist only in $\theta$ which is a relatively tiny network. On the other hand, the best performance reported in the distance-metric based approaches is the Prototypical Networks~\citep{snell2017prototypical} with $49.42\pm 0.78$. As a comparison, \model enjoys more flexibility without the need of defining any distance metric as in~\citep{vinyals2016matching,koch2015siamese,triantafillou2017few,snell2017prototypical,shyam2017attentive,mehrotra2017generative}.

\section{Related Work}
\label{sec:rela}
In this section, we provide brief discussions on related generative models in deep learning. In general, deep generative models can be categorized into two classes: explicit models and implicit models~\citep{goodfellow2016nips}. In explicit models, probability density or mass functions are explicitly defined, often parametrized with deep neural networks, so that the models can be learned or trained with statistical principles, e.g., maximum likelihood estimation or maximum a posteriori. Typical examples include variational autoencoder and its various variants~\citep{kingma2013auto}. As a comparison, implicit models do not provide a way for probability density/mass evaluation. Instead, the focus of implicit models lies in efficient sampling. This includes the popular generative adversarial networks~\citep{goodfellow2014generative} and its family. Without explicit density definitions, implicit models often rely on external properties of the model for training, including finding Nash equilibrium of the zero-sum game~\citep{goodfellow2014generative}, mixing distribution of the Markov chain~\citep{song2017nice,bordes2017learning}, etc. Using this taxonomy, our approach can be understood as an explicit generative model with well-defined probability density/mass functions. Compared with previous works in this line of research, \model has two significant improvements: first, existing works often assume training instances are $i.i.d.$ sampled from a fixed, probably unknown, distribution. On the other hand, we assume the data are sequentially generated from a Markov in an unknown order. Our model can thus discover the order and help us understand the implicit data relationships. Second, prior approaches were proposed based on the notion of denoising models. In other words, their goal was generating high-quality samples~\citep{kingma2013auto}. As a contrast, we aim at learning the dynamics of the chain whle at the same time recovering the order as well. 

\section{Conclusion}
In this technical report, we propose \model that can simultaneously learn the dynamics of a Markov chain and the generation order of data instances sampled from this chain. Our model is equipped with explicit, well-defined generative probabilities so that we can use statistical principles to train our model. Compared with previous work, \model uses flexible neural networks to parametrize the transitonal operator so that it amortizes the space complexity of the state space (only depends on the dimension of the state space, but not its cardinality), while at the same time achieves good generalization on unseen states during training. Experiments on discovering orders from unordered datasets validate the effectiveness of our model. As an application, we also extend our model to one-shot learning, showing that it performs comparably with state-of-the-art models, but with much less computational resources. 

\bibliography{reference}
\bibliographystyle{plainnat}

\end{document}


\maketitle

\section{Full Ordering Results for MNIST, Horse, and MSR-SenseCam}
\label{sec:full}

\begin{figure*}[t!]
	\centering
	\includegraphics[width=\textwidth]{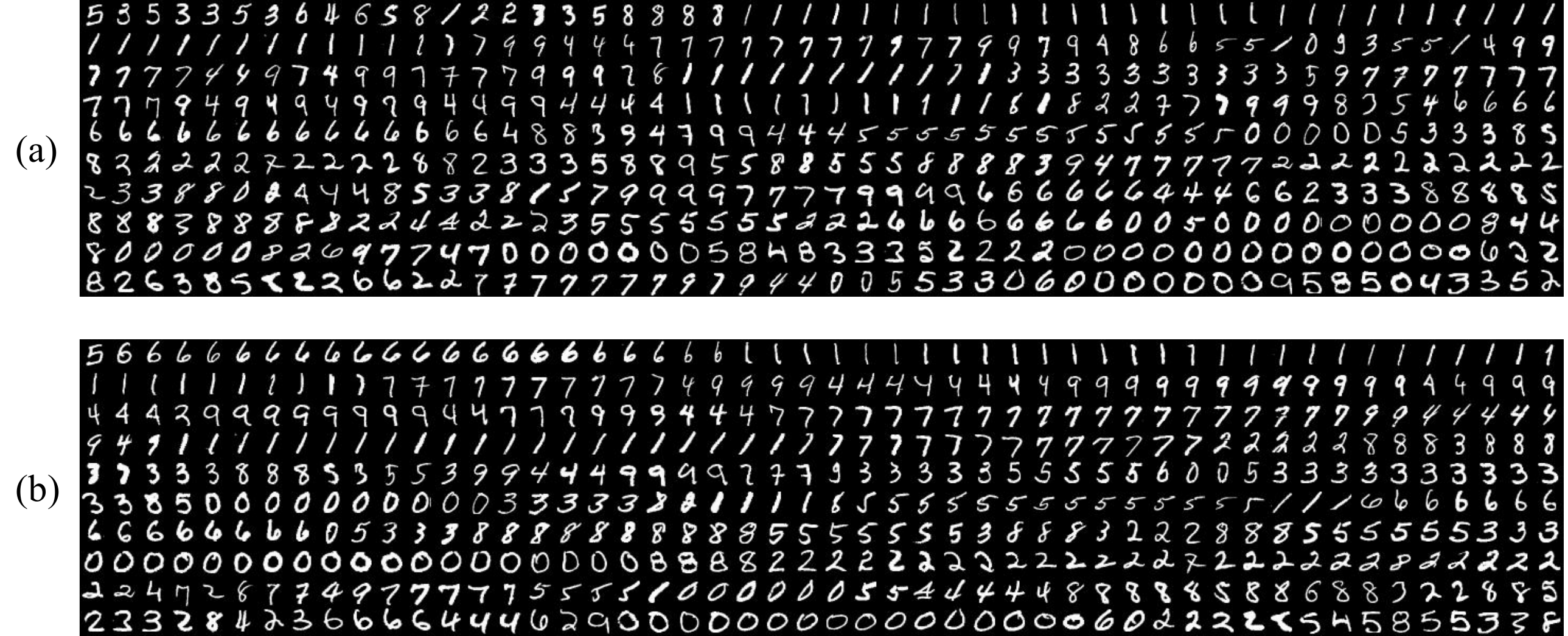}
	\caption{For MNIST dataset: (a) implicit order observed from OrderNet (b) order implied from nearest neighbor sorting using Euclidean distance.}
	\label{fig:mnist_impl_orders}
	\end{figure*}

\begin{figure*}[t!]
	\centering
	\includegraphics[width=\textwidth]{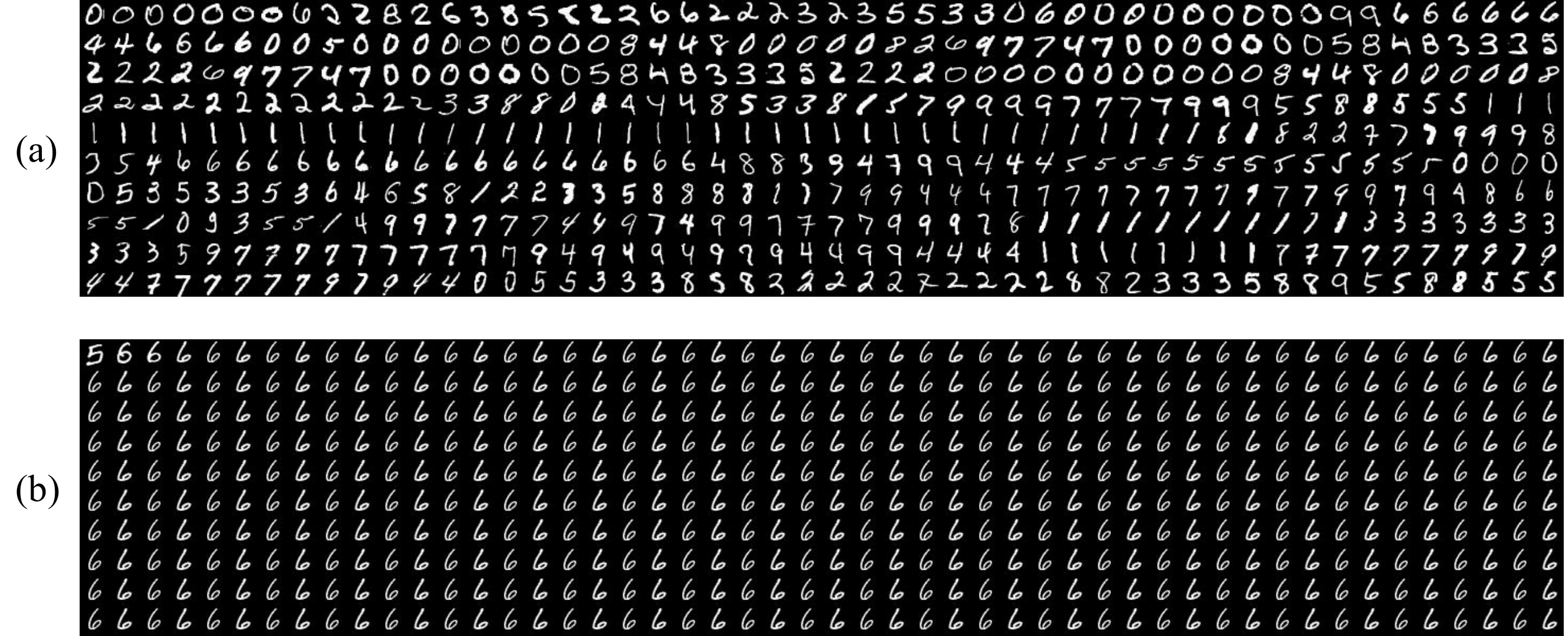}
	\caption{For MNIST dataset, {\em image propagation} from (a) learned transition operator in OrderNet (b) nearest neighbor search using Euclidean distance.}
	\label{fig:mnist_transition}
	\end{figure*}

\begin{figure*}[t!]
	\centering
	\includegraphics[width=\textwidth]{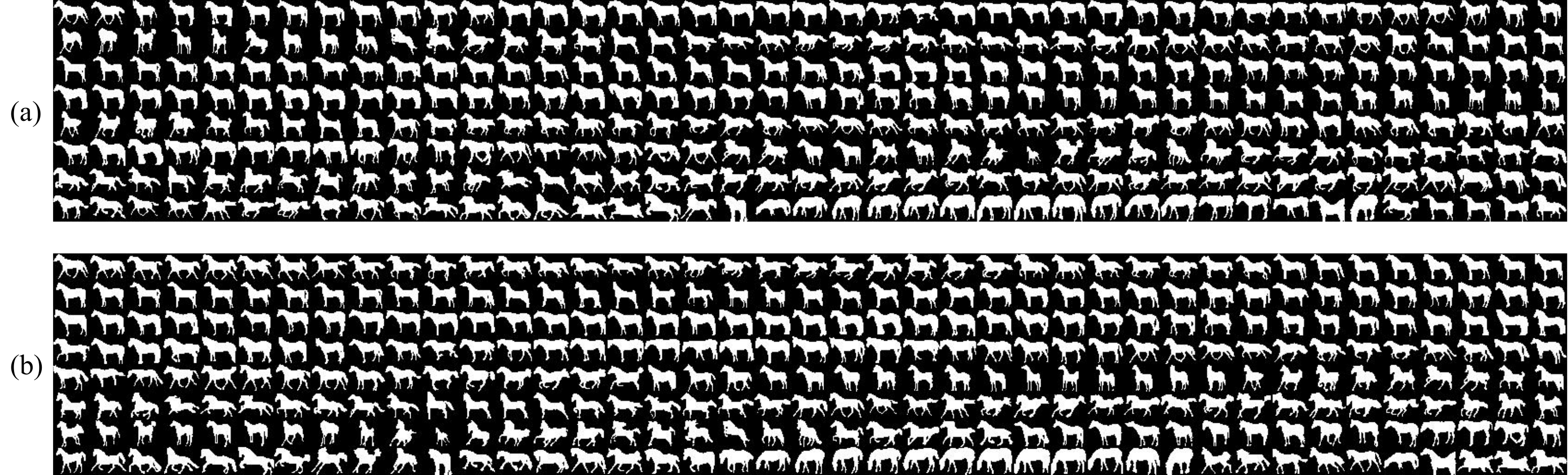}
	\caption{For Horse dataset: (a) implicit order observed from OrderNet (b) order implied from nearest neighbor sorting using Euclidean distance.}
	\label{fig:horse_impl_orders}
	\end{figure*}

\begin{figure*}[t!]
	\centering
	\includegraphics[width=\textwidth]{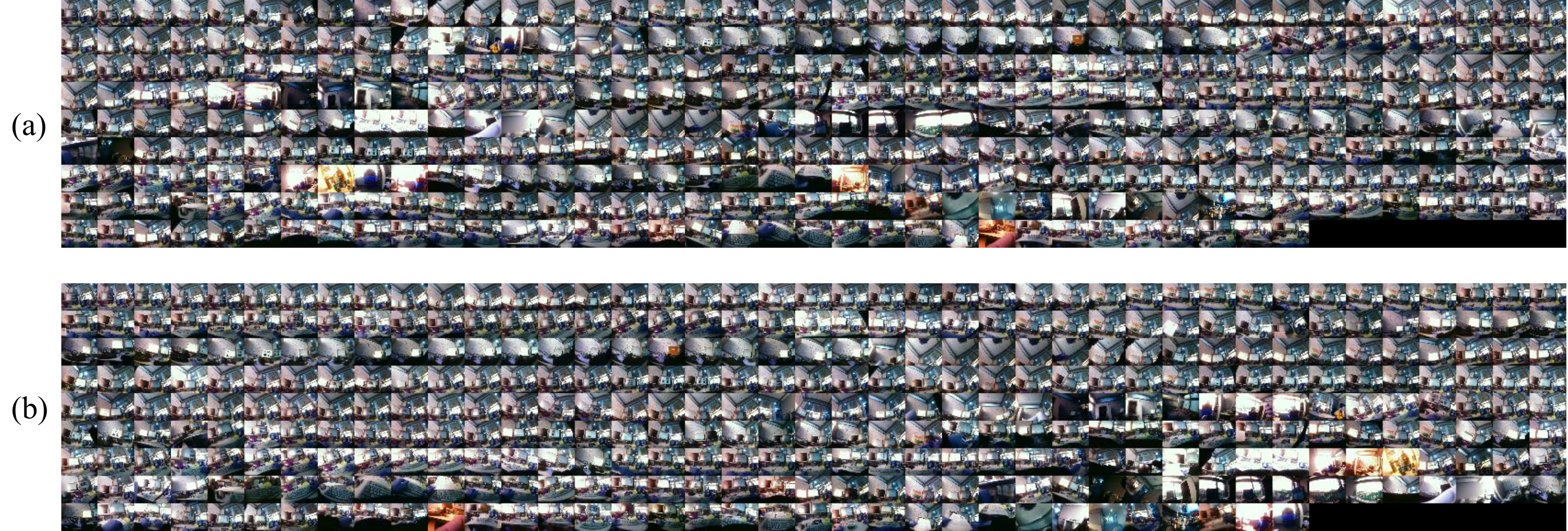}
	\caption{For {\em office} category in SenseCam dataset: (a) implicit order observed from OrderNet (b) order implied from nearest neighbor sorting using Euclidean distance.}
	\label{fig:sensecam_impl_orders}
	\end{figure*}

\begin{figure*}[t!]
	\centering
	\includegraphics[width=\textwidth]{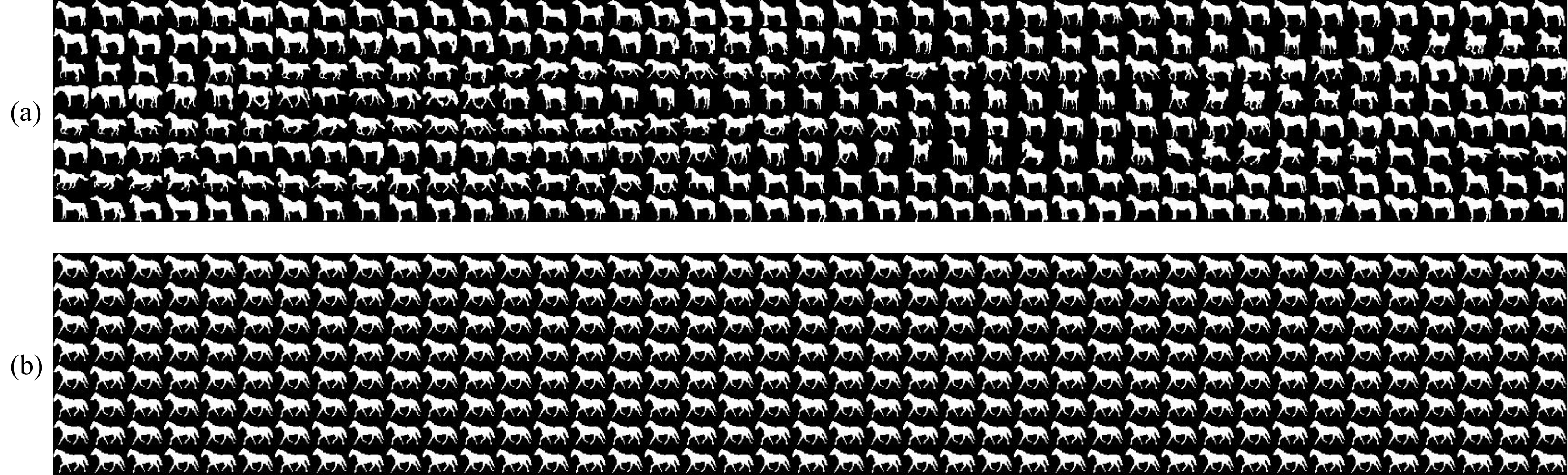}
	\caption{For Horse dataset, {\em image propagation} from (a) learned transition operator in OrderNet (b) nearest neighbor search using Euclidean distance.}
	\label{fig:horse_transition}
	\end{figure*}

\begin{figure*}[t!]
	\centering
	\includegraphics[width=\textwidth]{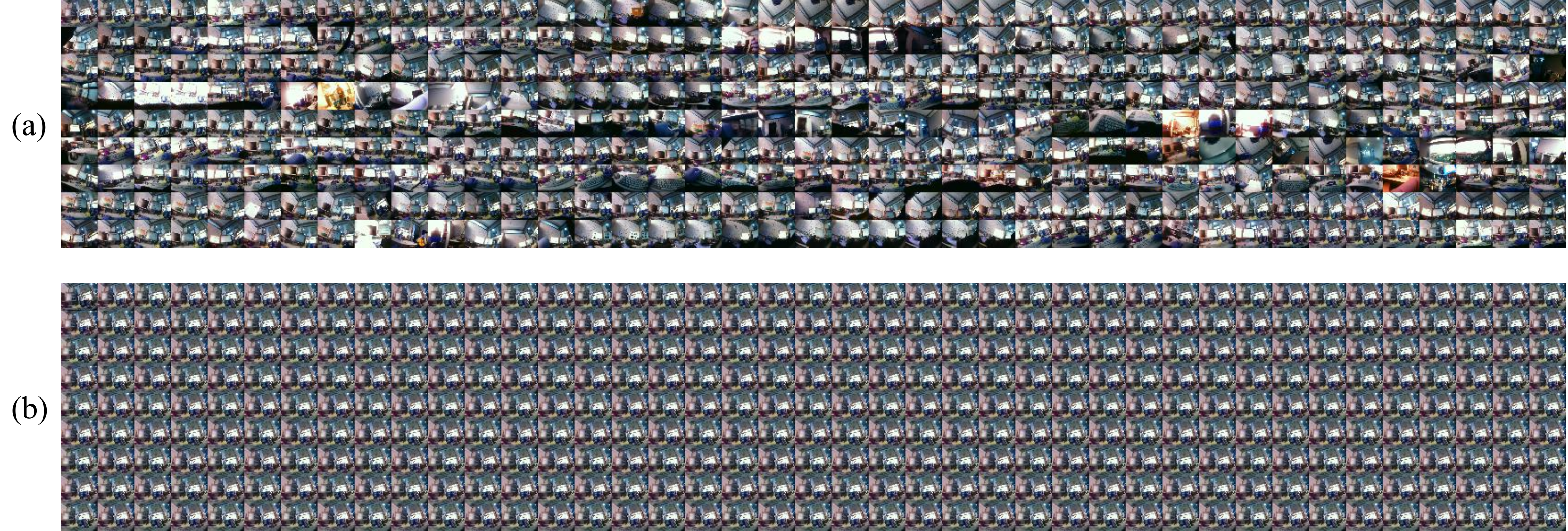}
	\caption{For {\em office} category in SenseCam dataset, {\em image propagation} from (a) learned transition operator in OrderNet (b) nearest neighbor search using Euclidean distance.}
	\label{fig:sensecam_transition}
	\end{figure*}

	Fig. \ref{fig:mnist_impl_orders}, \ref{fig:horse_impl_orders}, and \ref{fig:sensecam_impl_orders} show the results of the implicit order observed from OrderNet the order implied from Nearest Neighbor sorting. On the other hand, Fig. \ref{fig:mnist_transition}, \ref{fig:horse_transition}, and \ref{fig:sensecam_transition} illustrate the {\em image propagation} of OrderNet and Nearest Neighbor search. 

\section{Details for UCF-CIL Action Experiments}
\label{sec:ucf}

\begin{figure*}[t!]
	\centering
	\includegraphics[width=1.0\textwidth]{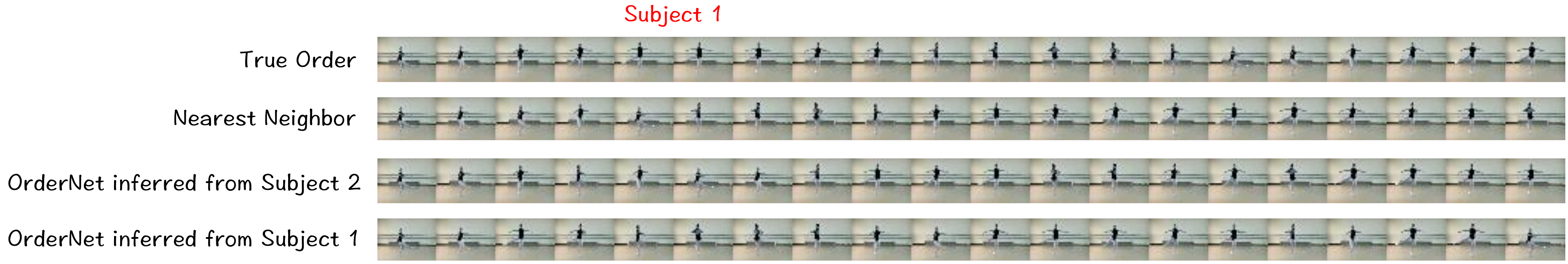}
	\caption{\small {\em Ballet fouette} action for $\mathsf{subject 1}$ with : (a) true order, (b) order recovered from Nearest Neighbor, (c) order recovered from OrderNet trained on {\bf different} subject, and (d) order recovered from OrderNet trained on the same subject.}
	\label{fig:recovery_full}
	\end{figure*}

	In the main text, we have provided the experiments on {\em ballet fouette} actions. Fig. \ref{fig:recovery_full} illustrates the comparison between (a) true order, (b) order recovered from Nearest Neighbor, (c) order recovered from OrderNet trained on {\bf different} subject, and (d) order recovered from OrderNet trained on the same subject. 

	Next, we provide the experiments on {\em tennis serve} actions. The results are provided in Tbl. \ref{tbl:tennis}. We can clearly see that, in most of the cases, the order implied by OrderNet enjoy better Kendall Tau-b values than Nearest Neighbor, which means our proposed model can recover more accurate orders.
	
\section{Hyper Parameters}
\label{sec:hyper}

    Tbl. \ref{tbl:hyper} lists the hyper parameters choice. Note that for smaller datasets (i.e., \textsf{Horse, MSR-SenseCam}), we can directly train OrderNet on entire dataset. In other words, $b_o = b$ and $t = 1$. Note that the number of frames is reduced to $20$ and $30$ for {\em ballet fouette} and {\em tennis serve} actions, respectively.  
    
\section{Network Architectures for Transition Operator}
\label{sec:transi_oper}

\begin{figure*}[t!]
	\centering
	\includegraphics[width=\textwidth]{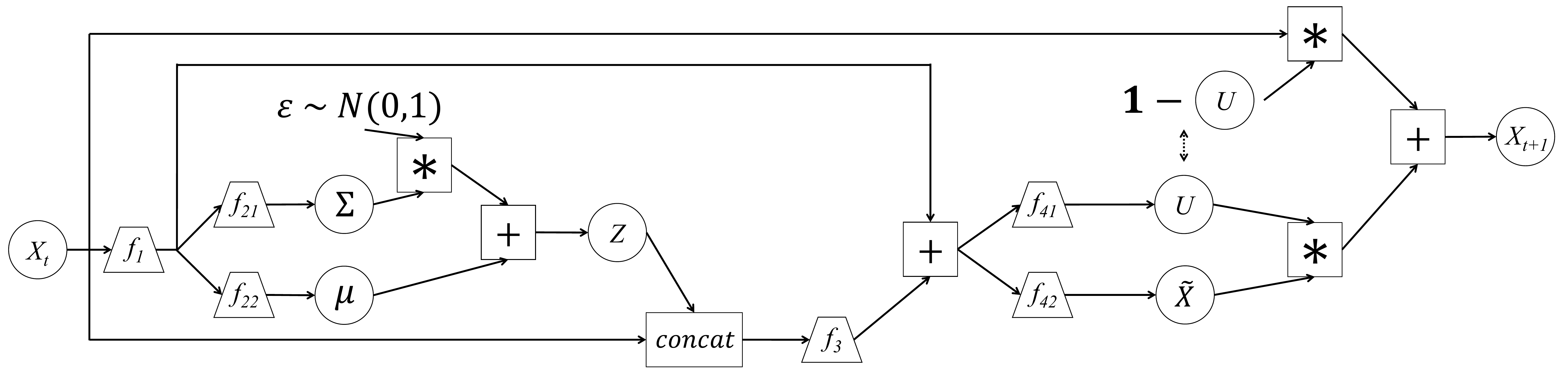}
	\caption{Network design for $\mathcal{T}(\cdot|\cdot; \theta)$.}
	\label{fig:transi}
	\end{figure*}

    	We elaborate the design of the transition operator in Fig. \ref{fig:transi}. In our design, $U$ can be seen as a gating mechanism between input $X_t$ and the learned update $\tilde{X}$. More precisely, the output can be written as
	$X_{t+1} = U \odot \tilde{X} + (\mathbf{1} - U) \odot X_t$,
	where $\odot$ denotes element-wise product. We specify each function $f$ in Tbl. \ref{tbl:horse_transi}, \ref{tbl:sensecam_transi}, \ref{tbl:mnist_transi},  \ref{tbl:ucf_transi}, and \ref{tbl:mini_transit}. Note that we omit the bias term for simplicity. 
	We use ADAM \citep{kingma2014adam} with learning rate $0.001$ and $0.2$ dropout rate to train our $\mathcal{T}(\cdot|\cdot;\theta)$.
	
\bibliography{reference}

\begin{table}[t!]
\centering
\caption{Hyper parameters choice.}
\label{tbl:hyper}
\vspace{-2mm}
\begin{tabular}{c||ccc}
\hline
\textsf{Hyper Parameters}          & $b_o$ & $b$ & $t$ \\ \hline \hline
\textsf{Horse}                     &     328     & 328  & 1  \\
\textsf{MSR\_SenseCam}               &    362      & 362  & 1  \\
\textsf{MNIST}                     &      50    &  500 &  600 \\
\textsf{UCF\_CIL} \textit{ballet fouette}          &  20    & 20   & 1  \\
\textsf{UCF\_CIL} \textit{tennis serve}      & 30    & 30 &  1 \\
\textsf{miniImageNet}              &   20   & 100  & 10  \\ \hline
\end{tabular}
\vspace{2mm}
\end{table}

\begin{table*}[t!]
\centering
\caption{\small Kendall Tau-b Metric between the true order and the recovered orders from OrderNet/ Nearest Neighbor (NN) with Euclidean distance. OrderNet are trained on different subjects with {\em tennis serve} actions for $10$ different subjects. The results are provided with mean and standard deviation from $20$ random trials.}
\vspace{2mm}
\label{tbl:tennis}
\scalebox{0.9}{
\begin{tabular}{cccccc}
\hline\hline
    & $\mathsf{Subject\,\,1}$     & $\mathsf{Subject\,\,2}$     & $\mathsf{Subject\,\,3}$    & $\mathsf{Subject\,\,4}$    & $\mathsf{Subject\,\,5}$     \\ \hline
$\mathsf{OrderNet}$ & -0.085 $\pm$ 0.034 & -0.147 $\pm$ 0.093 & 0.652 $\pm$ 0.032 & 0.377 $\pm$ 0.030 & 0.541 $\pm$ 0.048  \\
$\mathsf{NN}$  & 0.039         & -0.039        & 0.669        & 0.159        & 0.480           \\ \hline \hline
     & $\mathsf{Subject\,\,6}$    & $\mathsf{Subject\,\,7}$    & $\mathsf{Subject\,\,8}$    & $\mathsf{Subject\,\,9}$    & $\mathsf{Subject\,\,10}$    \\ \hline 
$\mathsf{OrderNet}$  & 0.034 $\pm$ 0.028 & 0.292 $\pm$ 0.065 & 0.518 $\pm$ 0.036 & 0.308 $\pm$ 0.038 & -0.343 $\pm$ 0.044 \\
$\mathsf{NN}$      & -0.172       & 0.209        & 0.370        & 0.126        & -0.163        \\ \hline \hline
\end{tabular}
}
\end{table*}

\begin{table*}[t!]
\centering
\caption{Details of functions for miniImageNet experiments.}
\vspace{2mm}
\label{tbl:mini_transit}
\begin{tabular}{cc}
\hline
function & details                       \\ \hline
f1       & 1600x1024 FC layer with ReLU //  1024x512 FC layer with ReLU // 512x256 FC layer with ReLU   \\
f21      & 256x64 FC layer              \\
f22      & 256x64 FC layer              \\
f3       & 1664x256 FC layer with ReLU    \\
f41      & 256x512 FC layer with ReLU // 512x1024 FC layer with ReLU // 1024x1600 FC layer with sigmoid \\
f42      & 256x512 FC layer with ReLU // 512x1024 FC layer with ReLU // 1024x1600 FC layer \\ \hline
\end{tabular}
\end{table*}

\begin{table}[t!]
\centering
\caption{Details of functions for Horse experiments.}
\label{tbl:horse_transi}
\vspace{-2mm}
\begin{tabular}{cc}
\hline
function & details                       \\ \hline
f1       & 1200x512 FC layer with ReLU    \\
f21      & 512x128 FC layer              \\
f22      & 512x128 FC layer              \\
f3       & 1328x512 FC layer with ReLU    \\
f41      & 512x1200 FC layer with sigmoid \\
f42      & 512x1200 FC layer with sigmoid \\ \hline
\end{tabular}
\vspace{2mm}
\end{table}

\begin{table}[t!]
\centering
\caption{Details of functions for MSR\_SenseCam experiments.}
\label{tbl:sensecam_transi}
\vspace{-2mm}
\begin{tabular}{cc}
\hline
function & details                       \\ \hline
f1       & 4096x1024 FC layer with ReLU    \\
f21      & 1024x256 FC layer              \\
f22      & 1024x256 FC layer              \\
f3       & 4352x1024 FC layer with ReLU    \\
f41      & 1024x4096 FC layer with sigmoid \\
f42      & 1024x4096 FC layer \\ \hline
\end{tabular}
\vspace{2mm}
\end{table}

\begin{table}[t!]
\centering
\caption{Details of functions for MNIST experiments.}
\label{tbl:mnist_transi}
\vspace{-2mm}
\begin{tabular}{cc}
\hline
function & details                       \\ \hline
f1       & 784x512 FC layer with ReLU    \\
f21      & 512x128 FC layer              \\
f22      & 512x128 FC layer              \\
f3       & 912x512 FC layer with ReLU    \\
f41      & 512x784 FC layer with sigmoid \\
f42      & 512x784 FC layer with sigmoid \\ \hline
\end{tabular}
\vspace{2mm}
\end{table}

\begin{table}[t!]
\centering
\caption{Details of functions for UCF\_CIL experiments.}
\label{tbl:ucf_transi}
\vspace{-2mm}
\begin{tabular}{cc}
\hline
function & details                       \\ \hline
f1       & 4118x1024 FC layer with ReLU    \\
f21      & 1024x64 FC layer              \\
f22      & 1024x64 FC layer              \\
f3       & 4182x1024 FC layer with ReLU    \\
f41      & 1024x4118 FC layer with sigmoid \\
f42      & 1024x4118 FC layer \\ \hline
\end{tabular}
\end{table}

\bibliographystyle{plainnat}